\documentclass[letterpaper, 10 pt, conference]{ieeeconf}  
\pdfoutput=1
\IEEEoverridecommandlockouts                               
\usepackage{cite}
\usepackage{amsmath,amssymb,amsfonts}
\usepackage{algorithmic}
\usepackage{graphicx}
\usepackage{textcomp}
\usepackage{xcolor}
\usepackage{subcaption}
\usepackage{caption}
\usepackage{multirow}
\usepackage{graphics}
\usepackage{epsfig}
\usepackage{times} 
\usepackage{amsmath} 
\usepackage{url}
\usepackage{color}
\usepackage{graphicx}
\usepackage{subcaption}
\usepackage{caption}
\usepackage{multirow}
\usepackage{url}
\usepackage{float}
\usepackage{sidecap}
\usepackage{tikz}
\usepackage[figurename=Fig.]{caption}
\newcommand\copyrighttext{
	\footnotesize \textcopyright 2023 IEEE. Personal use of this material is permitted.  Permission from IEEE must be obtained for all other uses, in any current or future media, including reprinting/republishing this material for advertising or promotional purposes, creating new collective works, for resale or redistribution to servers or lists, or reuse of any copyrighted component of this work in other works.
	DOI: 10.1109/SMC53992.2023.10394158}
\newcommand\copyrightnotice{
	\begin{tikzpicture}[remember picture,overlay]
		\node[anchor=south,yshift=10pt] at (current page.south) {\fbox{\parbox{\dimexpr\textwidth-\fboxsep-\fboxrule\relax}{\copyrighttext}}};
	\end{tikzpicture}
}

\title{\LARGE \bf
	Spatial and social situation-aware transformer-based trajectory prediction of autonomous systems
}

\author{Kathrin Donandt$^{1}$ and Dirk Söffker$^{2}$
	\thanks{$^{1}$Kathrin Donandt is with the Institute of Ship Technology, Ocean Engineering and Transport Systems, University of Duisburg-Essen, 47057 Duisburg,  Germany
		{\tt\small kathrin.donandt@uni-due.de}}%
	\thanks{$^{2}$Dirk Söffker is with the Chair of Dynamics and Control, University of Duisburg-Essen, 47057 Duisburg, Germany
		{\tt\small soeffker@uni-due.de}}%
}

\begin{document}
	\maketitle
	\copyrightnotice
	\thispagestyle{empty}
	\pagestyle{empty}

\begin{abstract}
Autonomous transportation systems such as road vehicles or vessels require the consideration of the static and dynamic environment to dislocate without collision. Anticipating the behavior of an agent in a given situation is required to adequately react to it in time. 
Developing deep learning-based models has become the dominant approach to motion prediction recently. The social environment is often considered through a CNN-LSTM-based sub-module processing a \textit{social tensor} that includes information of the past trajectory of surrounding agents. For the proposed transformer-based trajectory prediction model, an alternative, computationally more efficient social tensor definition and processing is suggested. It considers the interdependencies between target and surrounding agents at each time step directly instead of relying on information of last hidden LSTM states of individually processed agents. A transformer-based sub-module, the Social Tensor Transformer, is integrated into the overall prediction model. It is responsible for enriching the target agent's dislocation features with social interaction information obtained from the social tensor.  
For the awareness of spatial limitations, dislocation features are defined in relation to the navigable area. This replaces additional, computationally expensive map processing sub-modules. 
An ablation study shows, that for longer prediction horizons, the deviation of the predicted trajectory from the ground truth is lower compared to a spatially and socially agnostic model. Even if the performance gain from a spatial-only to a spatial and social context-sensitive model is small in terms of common error measures, by visualizing the results it can be shown that the proposed model in fact is able to predict reactions to surrounding agents and explicitely allows an interpretable behavior.
\end{abstract}

\begin{figure*}
	\centering
	\includegraphics[width=\textwidth]{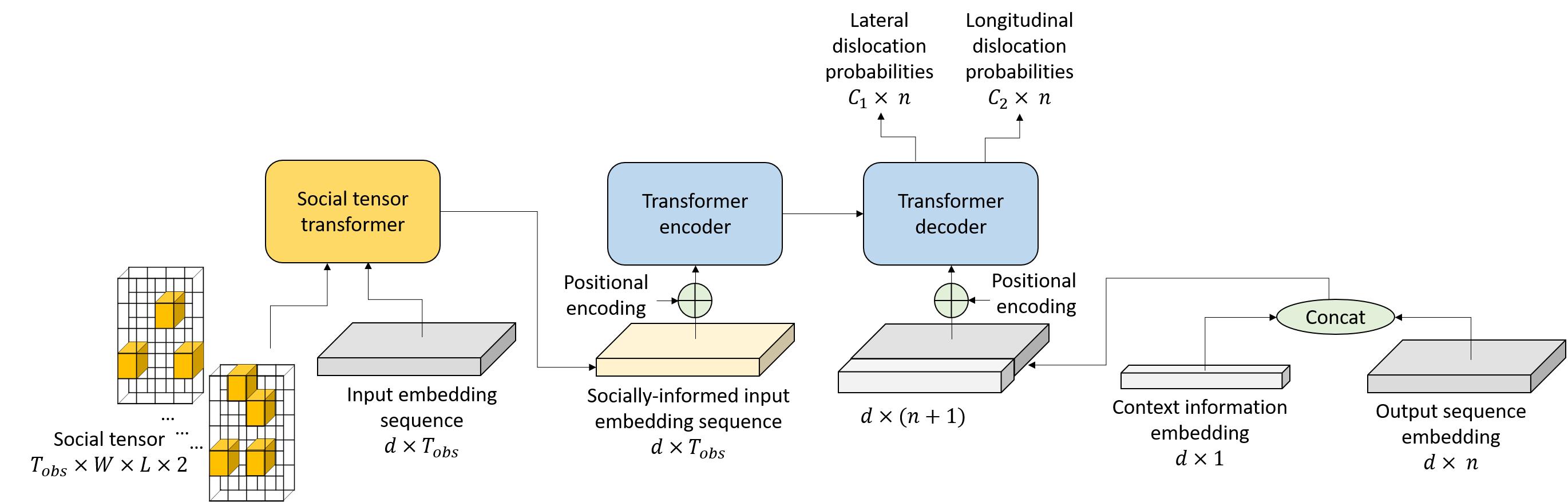}
		\caption{The sosp-CT model. Embedding layers are not depicted for better readability.
		The socially-informed embedding sequence obtained from the Social Tensor Transformer (see Fig. \ref{fig:stt}) and the concatenation of navigation context and output dislocations embedding are passed to the Transformer to generate the probability distributions over future dislocation labels. 
}\label{fig:modelarch}
\vspace{-1.5em}
\end{figure*}

\section{Introduction} 
In the context of intelligent and autonomous transportation systems, anticipating the behavior of traffic participants is an essential task to early detect dangerous situations and be able to avoid accidents by timely planned  collision-avoidance strategies. Behavior anticipation more specifically refers to the prediction of the spatio-temporal dislocation or trajectory in this contribution. 
The behavior of traffic agents depends not only on their own dynamics and physical possibilites, but constantly requires the reaction to and interaction with the static and dynamic environment, which is possible due to spatial and social situation awareness. The static environment refers to the space in which an agent is navigating, such as the road network with its lanes, crosswalks, etc., and the dynamic environment in the context of this study refers to other traffic participants, thus is also denominated the social environment. 
In analogy to Mozaffari et al. \cite{Mozaffari.2020}, the agent whose trajectory should be predicted is denominated the \textit{target agent} (TA). It differs from and should not be confused with the autonomous agent which uses the prediction model to be informed about possible future trajectories of TAs and thus to plan its future actions. This autonomous agent is referred to as the \textit{ego agent}. Finally, futher traffic participants which might influence the behavior of a TA are denoted \textit{surrounding agents} (SA). 
This contribution focusses on modeling the situation awareness of a TA for its trajectory prediction.
The developed prediction model can then be part of providing an intelligent, autonomous transportation system (the ego agent) with social situation awareness itself. 
Trajectory prediction is typically addressed by physics-based, statistical, traditional machine learning, or deep learning-based models. The latter have become increasingly popular due to their ability to accuratly predict trajectories even in complex traffic situations \cite{Lefevre.2014,Mozaffari.2020,Zhang.2022}. 
Recent deep learning-based models for trajectory prediction of a TA struggle to adequately model the interrelationship between the TA's behavior and that of the SA due to computational costs. The common usage of LSTM hidden state representation of the individual SA' trajectory histories fails to adequately  consider the interaction with the TA during the whole observation. 
Based on LSTM hidden states of the \textit{last} observed time step alone, the trajectory pediction model needs to reason about the interrelationship between the TA and the surrounding ones over the whole observation horizon. 
A further problem encountered in recently published approaches (e.g. \cite{Messaoud.2021,Li.2023,Chen.2022}), and also highlighted in \cite{Mozaffari.2020}, is the assumption of fully and continuously observable SA. This is questionable, as SA might enter the considered scene later, re-enter, or disappear from it. In this contribution, a new, transformer-based social tensor definition and processing considering the interrelationship at each time step between the trajectories of the TA and the fully or partially observed SA is introduced. In addition, an implicit spatial context consideration not requiring additional map data processing sub-modules is proposed. 

\section{Related Work}\label{sec:relwor}
Several survey papers report that deep learning-based methods are nowadays the state-of-the-art models for trajectory prediction for human, road vehicle, and vessel trajectory forecasting \cite{Rudenko.2020,Mozaffari.2020,Zhang.2022}. 
Target-centric forecasting methods try to predict the future motion of a single agent under consideration of its interaction with the static and dynamic environment. They are less complex and computationally expensive than models jointly anticipating the trajectories of multiple, interacting agents in a scene. 
To provide target-centric prediction models with spatial and social situation awareness, recent approaches often rely on CNN-based methods for the processing of map data and grid-representations of SA. 
Dijt and Mettes \cite{DijtPimandMettesPascal.2020} extract relevant information from the spatial environment by a YOLO model \cite{Redmon_2016_CVPR} that learns to semantically segment radar images by mapping them to Electronic Navigation Charts. This CNN is trained on the auxiliary task of semantic segmentation joinly with the RNN model responsible for predicting the future trajectory of inland vessels. Messaoud et al. \cite{Messaoud.2021} use rasterized, bird's-eye view maps of a region around the TA which includes information about lanes, driving directions, road shapes, drivable areas, side- and crosswalks. These maps are encoded by a pretrained ResNet-50 \cite{He.2016} to produce a feature representation that is later merged with a social tensor for a holistic context consideration. Li et al. \cite{Li.2023} guarantee scene-awareness by extracting features from semantic maps introduced in \cite{Houston.2020} which, in addition to geometries, include information about probable driving behaviors. The CNN-based feature extraction is handled by a pre-trained MobileNetV3 model \cite{Howard.2019}.  
Wang et al. \cite{Wang.2023} replace the CNN-based map processing by a more efficient transformer-based method taking as input vectorized maps. Roads are represented as a set of nodes in a directed graph and neighboring segments are connected by vectors containing coordinates, distances, road and segment identifiers. These vectorized maps are transformed into high-level features by a transformer encoder which are later fused with social context features. 
The SA' behavior and influence of it on the TA's motion is frequently handled by a combination of an LSTM and CNN. An occupancy grid around the TA, the social tensor, is populated with hidden states obtained when passing the past trajectory of each surrounding agent through an LSTM. In both Deo and Trivedi \cite{Deo_2018_CVPR_Workshops} and Li et al. \cite{Li.2023}, the social tensor is constructed from the last hidden LSTM states and then passed through a CNN. A high-level feature vector representation of the social context is obtained, which is concatenated to the spatial context feature vector to enable a situation-aware trajectory prediction. In Messaoud et al. \cite{Messaoud.2021}, the social tensor is constructed as in Deo and Trivedi \cite{Deo_2018_CVPR_Workshops}. It is then concatenated to the map representation and merged with the TA's trajectory representation by several attention heads separately to obtain a distribution over possible future trajectories. 
In Wang et al. \cite{Wang.2023}, the feature vectors of the SA' trajectory points are defined similarly to the map vectors described before, where distances are now calculated between subsequent points in time. A transformer encoder generates high-level features for each observed trajectory and fuses them in a multi-head attention block \cite{vaswani2017attention} with the features obtained from the map encoder. The fused context information is transformed into a feature vector that relates the TA's trajectory to the context through a further attention block. This feature vector is then passed to several transformer decoders which each predict a possible future trajectory of the target. Similarly to Wang et al. \cite{Wang.2023}, a transformer-based processing of the social environment is proposed in this contribution. An alternative method to generate the social tensor is suggested, which considers interdependencies between target and SA at each time step. For the spatial situation awareness, specific dislocation information definition is used instead of an additional model component responsible for the map processing.  
 
\section{Methodology}
Target-centric trajectory prediction can be considered a sequence-to-sequence modeling task where the upcoming positions of a moving agent are predicted based on observed positions. Additional information on the static and dynamic environment is required for a realistic outcome. Situation awareness is addressed here by using an extended version of the context-sensitive Classification Transformer model (CT) proposed in \cite{Donandt.2022}.

\subsection{Proposed model overview}
The overall model architecture of the proposed socially and spatially aware Classification Transformer, referred to as sosp-CT, is depicted in Fig. \ref{fig:modelarch}.   
Due to the reframing of the original regression task as a classification task (see \cite{Donandt.2022} for details), the  model is fed with sequences of discretized lateral and longitudinal dislocation information 
$(x_1,y_1),...,(x_{T_{obs}},y_{T_{obs}})$, where $(x_i, y_i)\in C_1\times C_2$ $\forall i \in [1;T_{obs}]$, $T_{obs}$ is the observation window, and $C_1$ and $C_2$ are the sets of lateral and longitudinal dislocation class labels, respectively. It predicts a probability distribution over the lateral and longitudinal dislocation labels for each prediction time step, and the future trajectory is reconstructed in a greedy manner by using the most likely dislocation labels at each time step. 
The lateral and longitudinal inputs are embedded each by a linear layer and concatenated to form the input embedding sequence of dimension $d \times T_{obs}$, where $d$ is the CT encoder and decoder input dimension. The output sequence fed during training to the decoder and the future dislocations obtained from the model predictions at inference are embedded analogously. The input embedding sequence is merged with the social tensor containing information of SA during observation by the Social Tensor Transformer (STT). 
Details on the definition of the social tensor and on STT are given in Section \ref{sec:sosa}. Apart from this model part responsible for the construction of the socially-informed input embedding sequence, the proposed model's architecture corresponds to that of the previous CT. 
The static or spatial context is only marginally considered by the CT through the usage of a specific decoder initialization. The CT decoder receives as very first input, i.e. when no previous target/predicted position is yet available, a context information embedding that encodes navigation area characteristics the TA is heading to.  
Additional spatial awareness is provided in the sosp-CT implicitly by making use of navigation-space related dislocation information as further explained in Section \ref{sec:spsa}. 
It is assumed that the longitudinal and transverse errors do not have the same contribution to the overall model and are therefore weighted differently in the loss function. The lateral and longitudinal cross entropy errors $\mathcal{L}_x$ and $\mathcal{L}_y$ are combined as proposed in \cite{Cipolla.2018} as
\begin{equation}
	\mathcal{L} = \frac{1}{\sigma_x^2}\mathcal{L}_x +  \frac{1}{\sigma_y^2}\mathcal{L}_y  + \log\sigma_x\sigma_y.
\end{equation}
The weights $\sigma_x$ and $\sigma_y$ are learnt jointly with the model parameters during training. 

\begin{figure}
	\centering
	\includegraphics[width=0.5\textwidth]{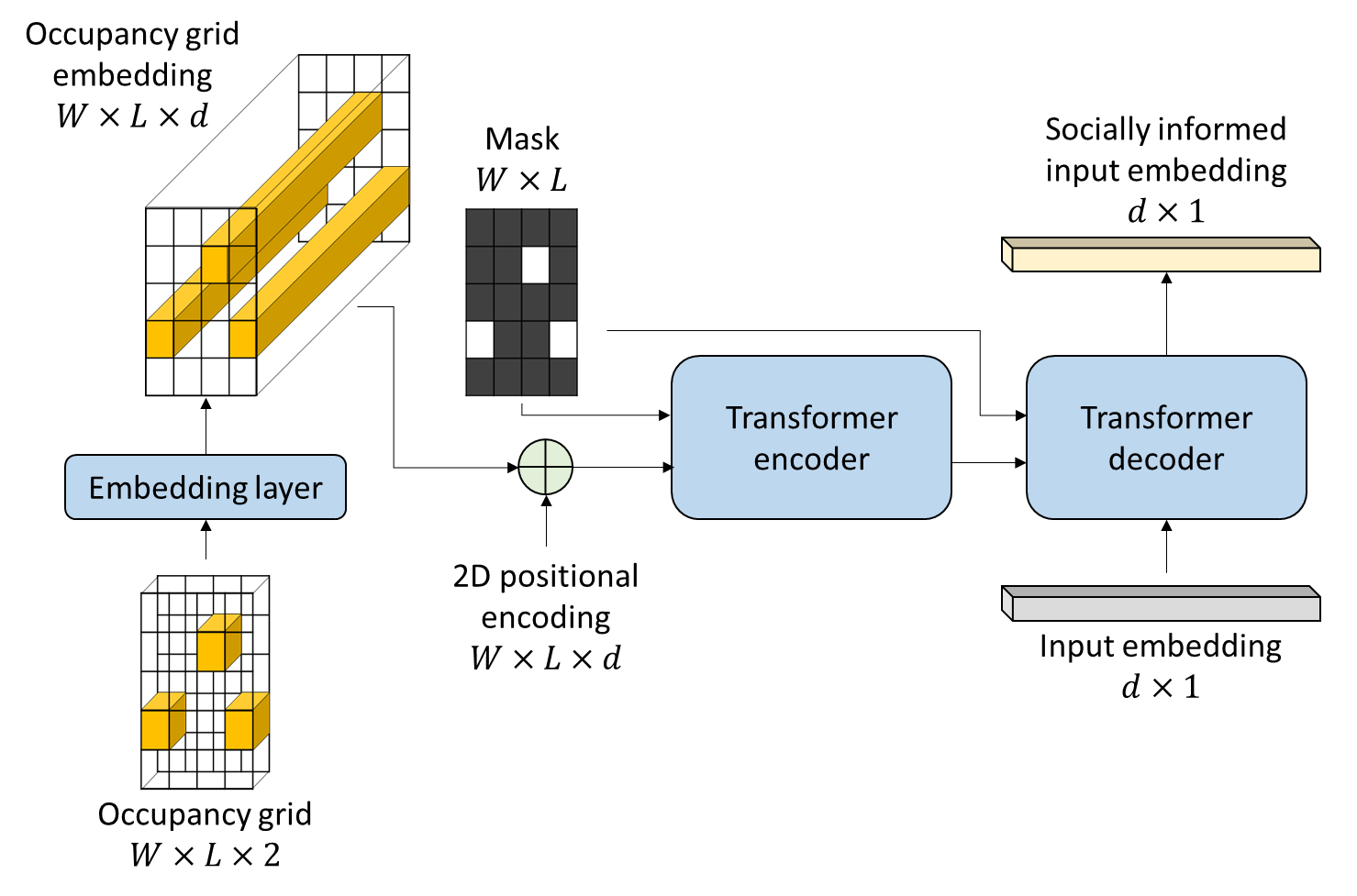}
	\caption{Social Tensor Transformer fusing social context and target agent dislocation features at each time step.} 
	\label{fig:stt}
\end{figure}
\begin{figure}
	\centering
	\begin{subfigure}{0.5\textwidth}
		\includegraphics[width=\textwidth]{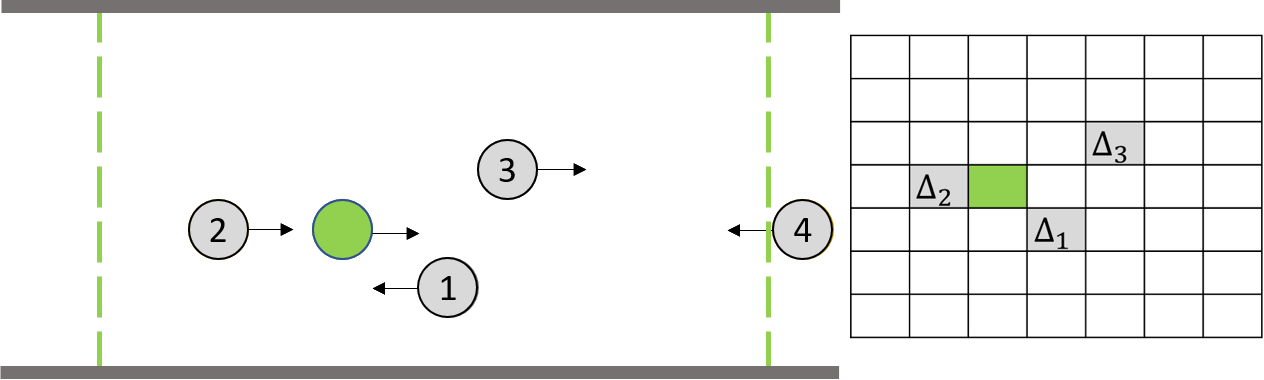}
		\caption{Time step 1}
		\label{fig:t1} 
		\vspace*{4mm}
	\end{subfigure}
	\vfill
	\begin{subfigure}{0.5\textwidth}
		\includegraphics[width=\textwidth]{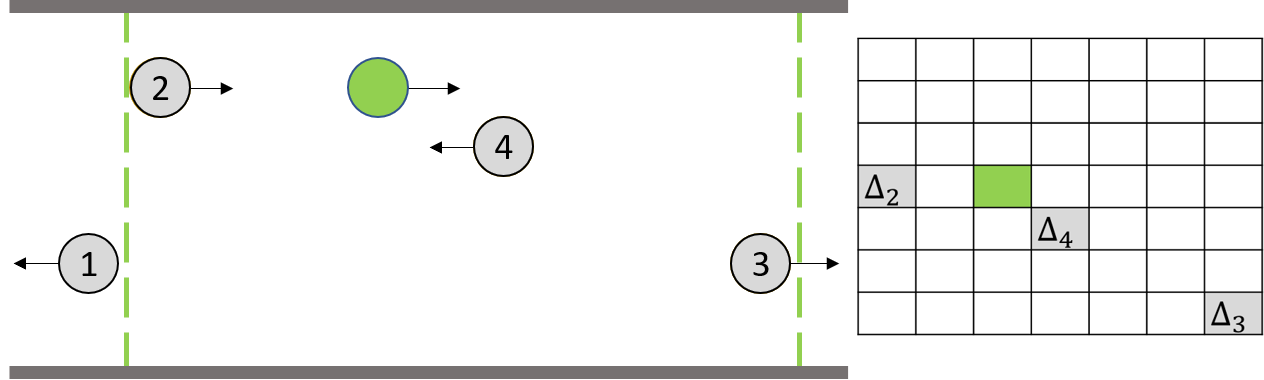}
		\caption{Time step 2}
		\label{fig:t2}
	\end{subfigure}
	\caption{Schematic example of traffic situations with occupancy grids. Dashed lines enclose the visual range of the target (green), dislocation change rates of surrounding agent $i$ are given by $\Delta_i$, and the target is included in the grid for better understanding.}\label{fig:og}
	\vspace{-1.5em}
\end{figure}
\subsection{Social situation awareness} \label{sec:sosa}
The area around the TA considered for social information inclusion and the agents which fall within it are commonly modelled by a social tensor in target-centric trajectory prediction approaches as mentioned in Section \ref{sec:relwor}. LSTM-based social tensors usually only contain the hidden state of the last observed time step, as computing the interrelationships at each time step would be computationally too costly \cite{Messaoud.2021}. 
Instead of LSTM hidden states, the cells of the occupancy grid around the TA are populated with the lateral and longitudinal distance change rates with respect to the TA directly here. The occupancy grids per observed time step are stacked to form the social tensor of the observation period. It is thus a 4D matrix of shape ($W$, $L$, $T_{obs}$, 2), where $W$ and $L$ are the grid dimensions. Each grid has its origin at the current position of the TA and is oriented into the direction of navigation.  
The vector at a specific grid position ($w$, $l$) of shape ($T_{obs}$, 2) covers different positions from a bird's-eye view, and thus might even by occupied by different agents, but correspond to one and the same position in a target-agent centric point of view. This is visualized in Fig. \ref{fig:og} for the case of agent 1 and 4. Surrounding agents are only considered in their relation to the TA at each time step. The knowledge of their individual dynamics, i. e. the temporal relationship between subsequent positions of each one of it - typically covered by the LSTM hidden state in common social tensor definitions - is considered to be irrelevant here.  
Due to the proposed construction of a social tensor at each observation time step, new, re-, or disappearing agents do not cause a modeling problem compared to social tensors constructions with LSTM hidden states of the last observed time step.
A social tensor element $s_t \in S$ at time step $t \in \lbrack 1;T_{obs}\rbrack$ is fused with the embedding of $\left( x_t,y_t \right)$ by the Social Tensor Transformer (STT) as depicted in Fig. \ref{fig:stt}. 
First, $s_t$ is transformed into a $W \times L \times d$ feature matrix by an embedding layer. To each position of the occupancy grid embedding, a 2D positional encoding \cite{Wang.2021} is added to provide the model with spatial location information of each grid position. A mask is additionally required to inform the model about which positions are unoccupied in the occupancy grid and thus should not be attended to. The positionally encoded occupancy grid embedding and the mask are then fed to a transformer encoder. The obtained feature representation and the mask are further passed to the transformer decoder which takes as input only the input embedding of the TA at time step $t$. 
In the decoder, the attention of the TA input embedding at that time step with respect to the non-masked, encoded occupancy grid positions obtained from the encoder is, in analogy to \cite{vaswani2017attention}, calculated as 
\begin{equation}
	softmax \left( \frac{qW^q(KW^K)^T}{ \sqrt{d} } \right) VW^V ,
\end{equation}
where the query $q$ is the target input embedding, both keys $K$ and values $V$ correspond to the non-masked elements of the encoded occupancy grid, and $W^q$, $W^K$, and $W^V$ are learnable parameter matrices. Through the decoder's attention mechanism, the model learns to weight the SA at time step $t$ according to their relevance for the TA's state at this time. This way, a socially-informed input embedding is obtained from the STT, which is trained jointly with the main CT model.   

\subsection{Spatial situation awareness}\label{sec:spsa}
\begin{SCfigure}
	\centering
		\includegraphics[width=0.228\textwidth]{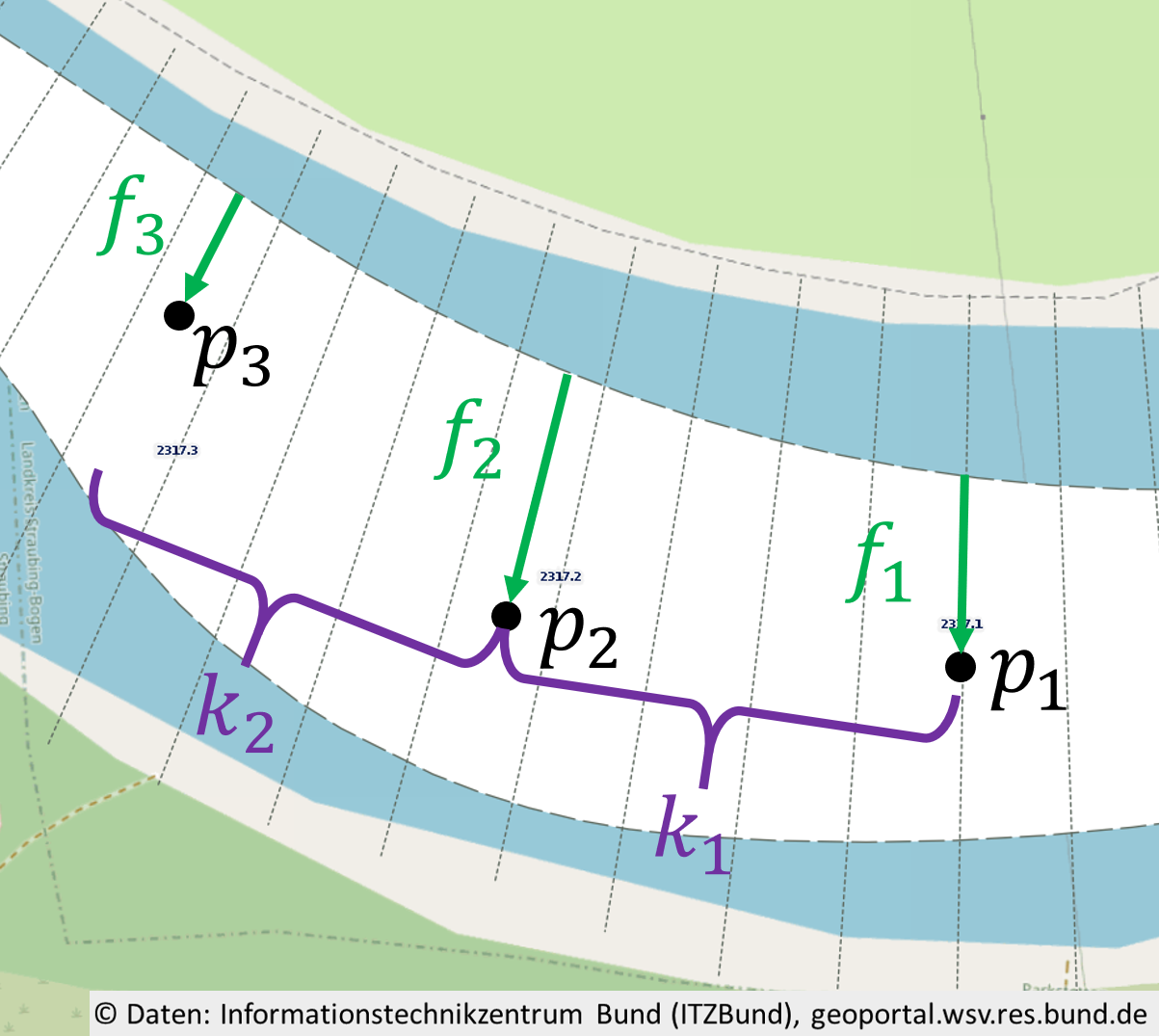}
	\caption{Navigation area-specific (dis)location in- formation. The white area depicts the fairway. Here, $k_i$ is the waterway kilometer distance between the positions $p_{i+1}$ and $p_i$ and $f_i$ the distance from the fairway border.}
	\label{fig:spa}
\end{SCfigure}

Different from previous approaches, an additional sub-module for handling map data is not required for spatial situation awareness here. Instead, the lateral and longitudinal dislocation of the TA is defined in relation to the navigable area and neither to the global reference system nor to a TA position at a specific time step. Fig. \ref{fig:spa} exemplifies this for the inland vessel scenario. 
The lateral and longitudinal dislocation is given by the change of the relative distance of the TA to the right fairway border and the distance of the waterway kilometers of the TA's coordinates between subsequent time steps. It should be noted that the occupancy grid dimensions $W$ and $L$ also make use of the navigation-area related coordinate system. In the given use case, a grid position thus corresponds to a waterway kilometer distance from the target vessel and a difference in the relative fairway border distance. 
 The dislocation information vector can be extended to higher dimensions if further information is required. 

\section{Use case: Inland vessel trajectory prediction}

The effectiveness of the proposed spatial and social situation-aware transformer-based prediction model is analyzed in the context of future inland vessel trajectory anticipation on a highly-navigated waterway. 

\subsection{Data source} 
The inland vessel trajectory prediction research cannot rely on the abundancy of benchmark datasets available e.g. for pedestrian and road vehicle trajectory prediction. For the chosen use case, a dataset containing relevant target vessel sequences to train the sosp-CT model needs to be constructed first. Vessel navigation data is logged with the Automatic Identification System (AIS) with a frequency typically around 0.1 Hz.  
An AIS dataset covering 2 years of selected Rhine sections is used. Regions near harbors and ferry routes are not included as traffic situations there are considered too specific, requiring to be handled separately. The AIS data of each day and river section are grouped by the Maritime Mobile Service Identities which uniquely identify a vessel. An inland vessel's trip is defined here as having a single navigation direction (up- or downstream) and a continuous AIS signal. The data of the vessel are therefore split into distinct trips at navigation direction changes or signal loss of more than 60 min. Different from target vessels which are only considered during navigation, surrounding vessel trips are allowed to include pauses at moorings, as moored vessels might also influence the behavior of target vessels passing nearby. 
Thresholds obtained by a statistical evaluation are used to exclude outliers with exceptionally high speed or acceleration. Cubic Hermite spline interpolation which considers the velocities is applied to UTM coordinate sequences of the trips to generate smooth and regularly sampled trajectory data. Interpolation is only applied between time intervals of less than 2 min.  
Waterway kilometers and distances to the fairway border are added. 
Trips are merged on the time step and TA sequences of 9 or 10 min, depending on the time resolution, together with relevant surrounding vessels' data are extracted. Both $T_{obs}$ and $n$ are identical, and either 4.5 min (90 s resolution) or 5 min (30 s or 60 s resolution). As driving behavior on rivers differs by navigation direction due to the current, the development of a direction-specific model is prefered. In this use case, an upstream model is developed and thus only upriver navigating target vessels are considered. Distance thresholds of 1.5 waterway kilometers ahead of the target vessel and 0.75 behind are applied to distinguish between surrounding and non-effective vessels. Encountering vessels navigate downstream and thus typically traverse a longer river section compared to vessels approaching from behind in a given time horizon. 
Target vessel sequences assumed relevant are obtained by making sure that at least one encounter or overtaking takes place during the prediction horizon. The final training dataset contains $\approx$400k sequences. 

\subsection{Results}
\begin{figure}
	\begin{subfigure}{0.48\textwidth}
		\includegraphics[width=\textwidth, height=0.215\textheight]{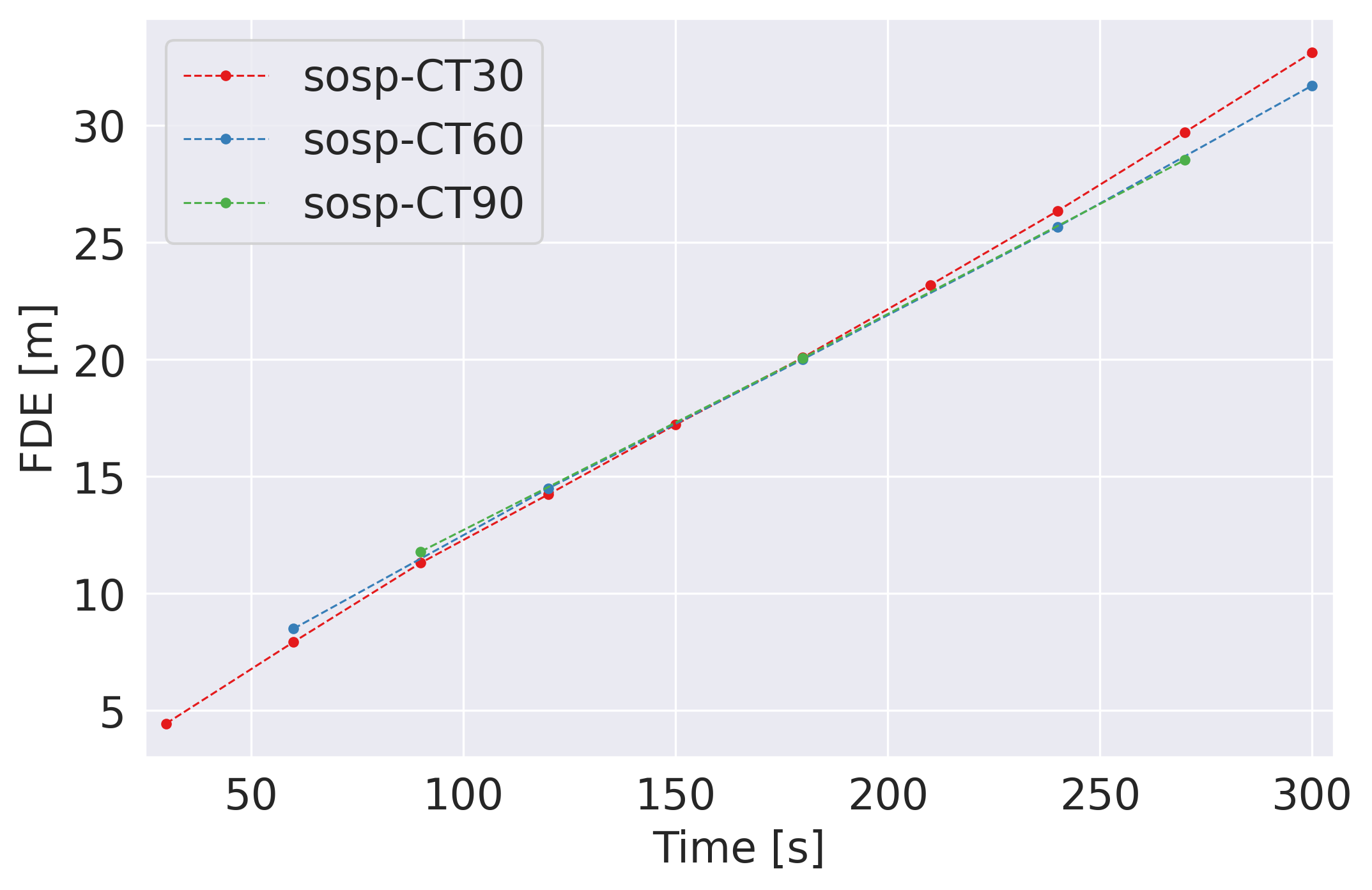}
		\caption{Comparison of time resolutions.}
		\vspace*{4mm}
		\label{fig:ts}
	\end{subfigure}
	\begin{subfigure}{0.48\textwidth}
		\includegraphics[width=\textwidth, height=0.215\textheight]{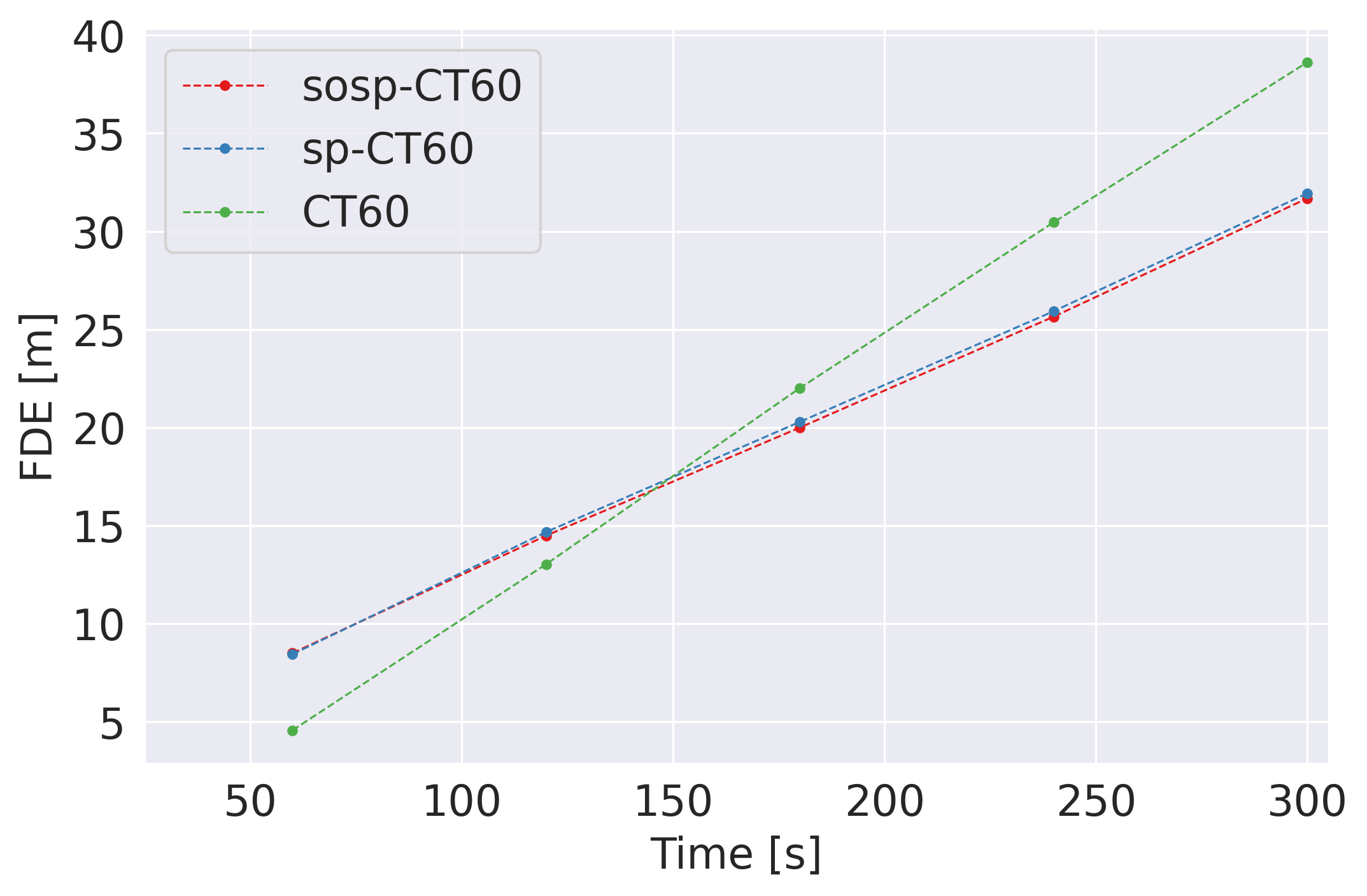}
		\caption{Ablation study results.}
		\label{fig:sosp_sp}
	\end{subfigure}	
	\begin{subfigure}{0.5\textwidth}
		\includegraphics[width=\textwidth,height=0.24\textheight]{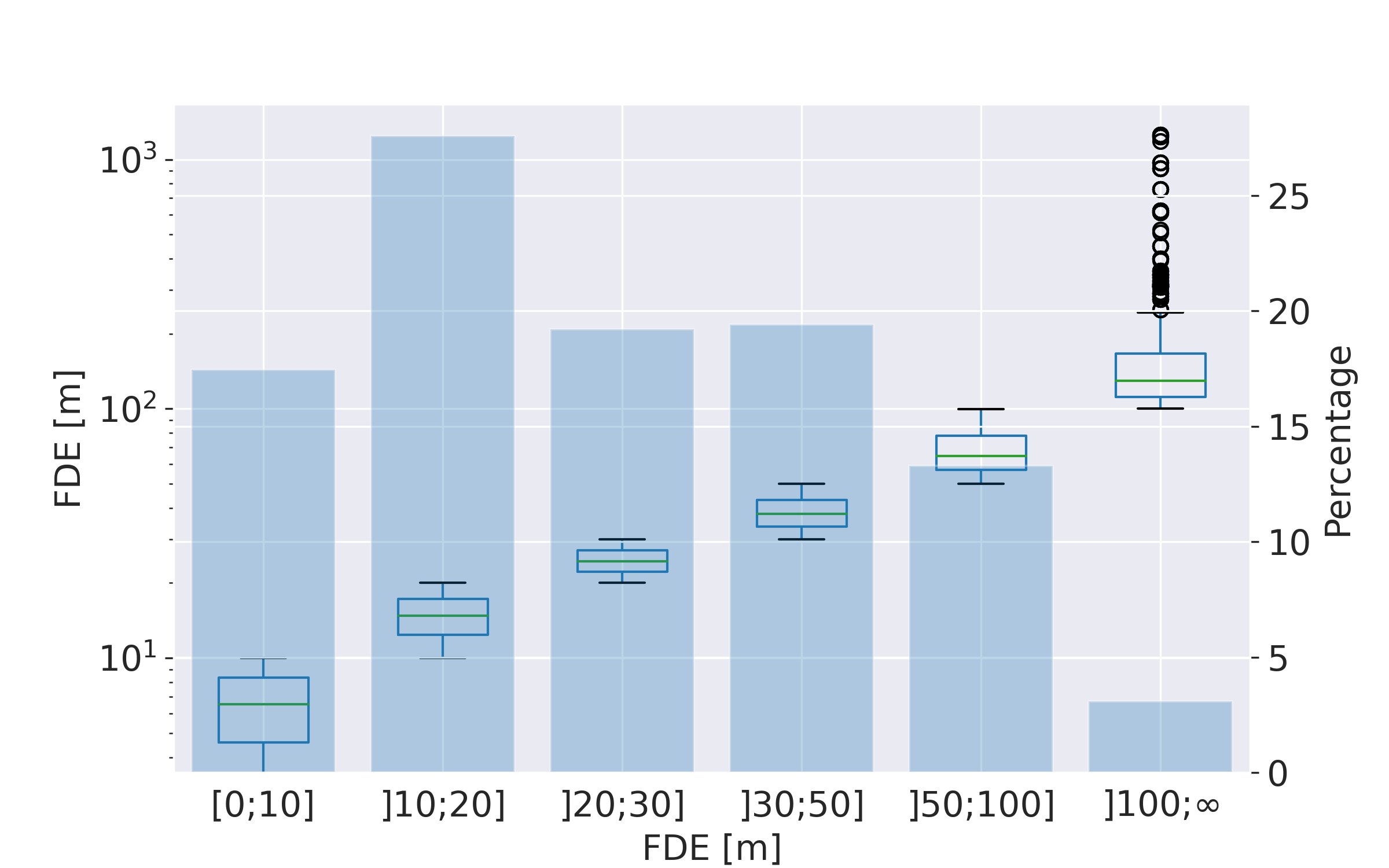}
		\caption{FDE distribution of sosp-CT60.}
		\vspace*{2mm}
		\label{fig:fdedistpc}
	\end{subfigure}
	\caption{FDE evolution and distribution.}
\vspace{-1.5em}
\end{figure}
An ablation study is carried out to analyze the benefits of providing a trajectory prediction model with social and spatial awareness. The common metrics average and final displacement error (ADE \& FDE) are used for evaluation. 
The previous model, CT, is compared to a model with spatial awareness only, sp-CT, and the proposed sosp-CT model with both contexts considered. Different time resolutions are compared (30, 60, and 90 s). For the 30 and 60 s resolution, the model is trained to predict 5 min, and for the 90 s resolution 4.5 min. The FDE is plotted with different time steps considered `final' in Fig. \ref{fig:ts} and Fig. \ref{fig:sosp_sp}. The context-agnostic model is better for predictions less than 3 min, which can probably be attributed to the small discrepancies between waterway kilometer distances and real distances. This causes small inconsistencies in the navigation area-related dislocation features. Also, vessel behavior does not change rapidly due to the vessels' inertia. This could be a further reason why the context-agnostic model is not outperformed by the proposed model for small time horizons. This hypothesis is further confirmed by the time resolution comparison. A model trained on a higher resolution shows weaker performance. Behavior changes might be too small in consecutive small time steps and therefore, such a model struggles to learn behavior adaptations clearly noticable only after several time steps.  
The ADE and FDE for the whole prediction horizon for a time step size of 60 seconds is given in Table \ref{tab:adefde}.
\setlength{\tabcolsep}{0.5em}
{\renewcommand{\arraystretch}{1.2}
	\begin{table}[]
		\centering
		\caption{ADE and FDE after 5 min (60 s resolution)}
		\begin{tabular}{|c||c|c|}
			\hline
			\textbf{Model}   & \textbf{ADE} & \textbf{FDE} \\ \hline 
			\hline
			\textbf{CT} & 21.70 $\pm$ 28.2 & 38.55 $\pm$ 46.8 \\ \hline
			\textbf{sp-CT} & 20.25 $\pm$ 26.5 & 31.95 $\pm$ 44.32 \\ \hline
			\textbf{sosp-CT} & \textbf{20.06 $\pm$ 25.11} & \textbf{31.69 $\pm$ 40.58} \\ \hline
		\end{tabular}\label{tab:adefde}
	\end{table}The best performance is obtained by sosp-CT. The differences to the sp-CT model is, however, small. 
When analyzing the large number of errors, it is noticed that these are trajectories with an unusual behavior (very fast for an upstream vessel). In Fig. \ref{fig:fdedistpc}, the error distribution shows that more than 80\% of the test samples have an error less or equal to 50 m, and less than 4\% have an error higher than 100 m after 5 min. The few outliers contribute to the high standard deviation shown in Table \ref{tab:adefde} and Fig. \ref{fig:fdedistpc}. 
Apart from the small numerical difference in FDE between sp-CT and sosp-CT, cases can be identified, in which the sosp-CT is able to predict correct reactions to a given traffic situation and the sp-CT is not. An example is shown in Fig. \ref{fig:ex}, where the sosp-CT model correctly predicts the maneuver back to the target vessel's original path after having sidestepped due to the encounter with the blue-colored  
vessel. The sp-CT model erroneously maintains the behavior observed shortly before. 
This example indicates the potential of sosp-CT. Further model optimizations and data improvements are required to make the advantage of sosp-CT over sp-CT and CT more visible. 
\begin{figure}
		\centering
		\begin{subfigure}{0.5\textwidth}
				\includegraphics[width=\textwidth]{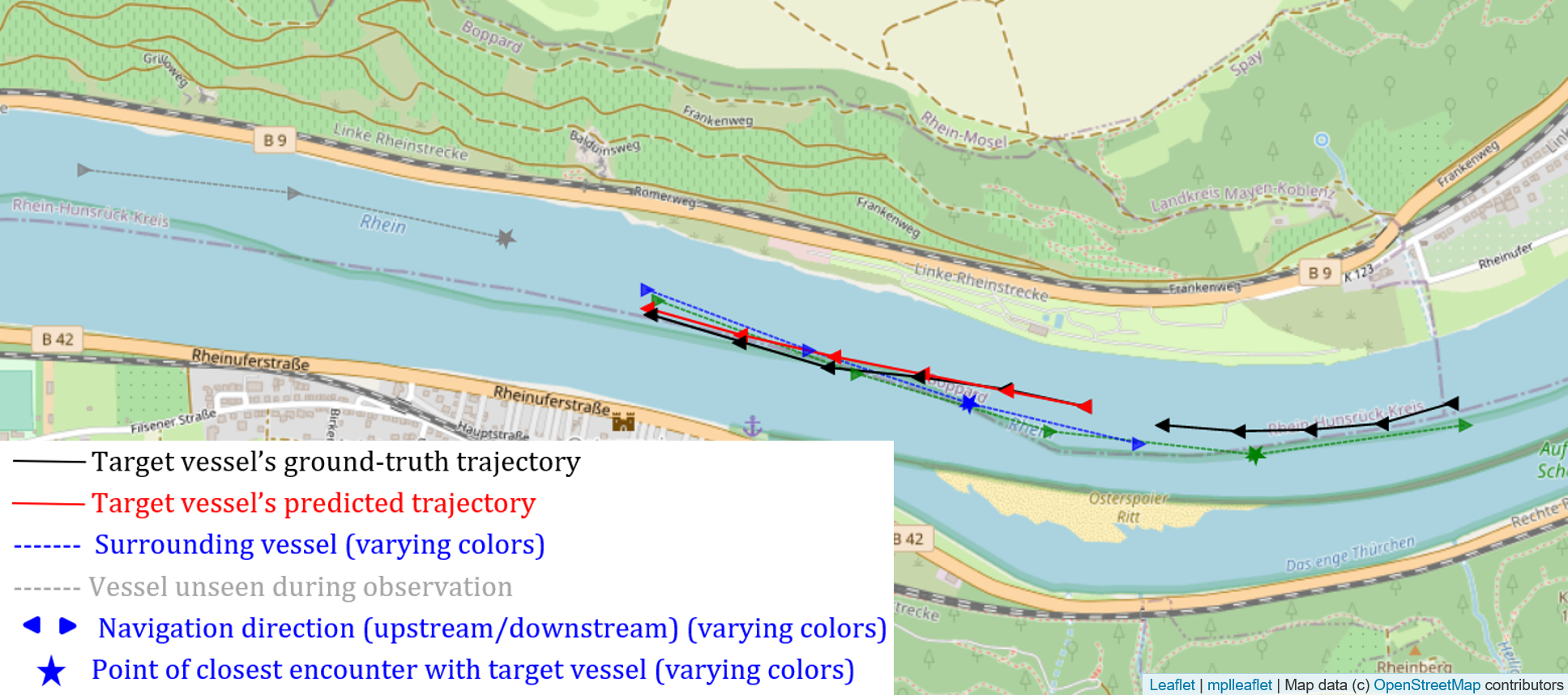}
				\caption{sosp-CT60}
				\vspace*{4mm}
				\label{fig:3D}
			\end{subfigure}
		\begin{subfigure}{0.5\textwidth}
				\includegraphics[width=\textwidth]{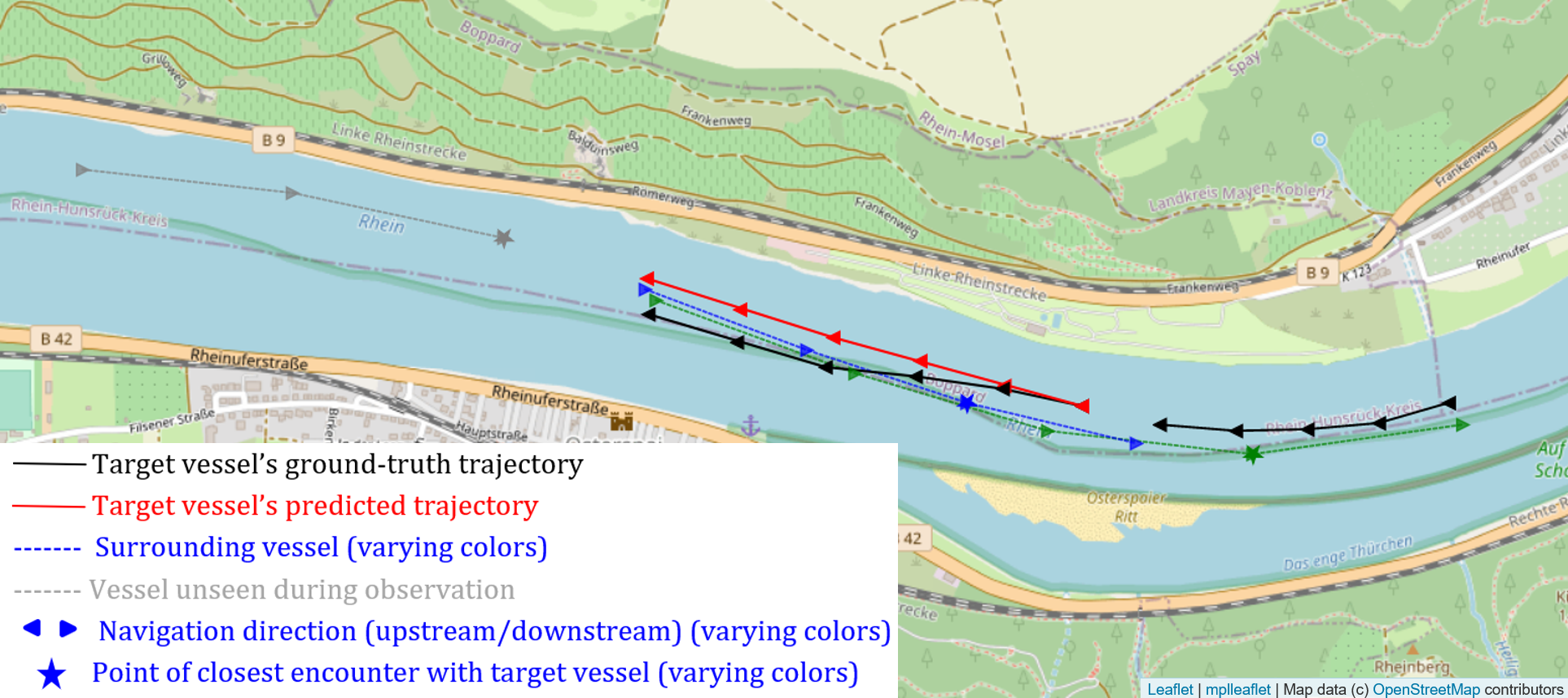}
				\caption{sp-CT60}
				\label{fig:2D}
			\end{subfigure}
		\caption{Prediction example of the socially- and spatially-aware (a) and the spatially-aware model (b).}\label{fig:ex}
\vspace{-1.5em}
	\end{figure}
\section{Summary, Conclusion \& Outlook}
A new method for the consideration of the social and spatial context in a deep learning-based target-specific trajectory prediction model is presented. Spatial awareness is ensured by defining the model input features in relation to the navigation area, and no additional map processing sub-module is required to inform the model about the navigable space. The prediction model is informed about surrounding agents during observation at each time step. This way, the interrelation between the target agent's trajectory with the surrounding traffic is considered over the whole observation horizon explicitly. This is different from previous approaches, which instead rely on the model's capacity to extract this interrelationship information from individually encoded trajectory histories. 
The method is tested for the use case of inland vessel trajectory prediction. It can be shown that the consideration of spatial and social information results in better predictions. Especially the navigation area-specific input representation results in a lower final displacement error compared to a situation-agnostic model.
The model is computationally more efficient than previous models relying on LSTM-based social tensors and on map inputs as it uses a fully transformer-based architecture and makes map processing dispensable. The model is able to deal with partially observed surrounding agents, which is important as signal failures can occur and agents are not guaranteed to be present in the considered scene during the entire observation window. In future works, a comparison of the prediction performances of the proposed transformer-based model and previously suggested CNN-LSTM-based methods is intended. 
An extensive hyperparameter tuning, and further data filtering are expected to improve the results obtained so far.  
Further experiments on existing benchmark datasets for pedestrian and road vehicle trajectory prediction are planned to prove the method's generalizability to other use cases.

\section*{Acknowledgment}
The authors thank the German Federal Waterways Engineering and Research Institute for the provision of AIS and river-specific data, the trip splitting and kilometerization algorithms, and the training infrastructure. 

\bibliographystyle{IEEEtran}
\bibliography{bib}

\end{document}